\documentclass[sigconf]{acmart}
\usepackage{cleveref}
\usepackage{subfigure}
\usepackage{diagbox}
\usepackage{multirow}
\usepackage{relsize}
\newcommand{\indep}{\rotatebox[origin=c]{90}{$\models$}}
\begin{document}

\fancyhead{}
\settopmatter{printacmref=true}
\title{Coupled Variational Recurrent Collaborative Filtering}

\author{Qingquan Song$^1$, Shiyu Chang$^2$, Xia Hu$^1$}
\affiliation{%
  \institution{$^1$Department of Computer Science and Engineering, Texas A\&M University} 
  \institution{$^2$MIT-IBM Watson AI Lab, IBM Research}
}
\email{{song_3134, xiahu}@tamu.edu, shiyu.chang@ibm.com}


\renewcommand{\shortauthors}{Qingquan Song, et al.}

%
\begin{abstract}

We focus on the problem of streaming recommender system and explore novel collaborative filtering algorithms to handle the data dynamicity and complexity in a streaming manner. Although deep neural networks have demonstrated the effectiveness of recommendation tasks, it is lack of explorations on integrating probabilistic models and deep architectures under streaming recommendation settings.  Conjoining the complementary advantages of probabilistic models and deep neural networks could enhance both model effectiveness and the understanding of inference uncertainties.  To bridge the gap, in this paper, we propose a \underline{C}oupled \underline{V}ariational \underline{R}ecurrent \underline{C}ollaborative \underline{F}iltering (CVRCF) framework based on the idea of Deep Bayesian Learning to handle the streaming recommendation problem. The framework jointly combines stochastic processes and deep factorization models under a Bayesian paradigm to model the generation and evolution of users' preferences and items' popularities.  To ensure efficient optimization and streaming update, we further propose a sequential variational inference algorithm based on a cross variational recurrent neural network structure. Experimental results on three benchmark datasets demonstrate that the proposed framework performs favorably against the state-of-the-art methods in terms of both temporal dependency modeling and predictive accuracy. The learned latent variables also provide visualized interpretations for the evolution of temporal dynamics.
\end{abstract}

%
%
\keywords{collaborative filtering, streaming recommender system, matrix factorization, deep Bayesian learning}

\maketitle

\section{Introduction}

With the explosive growth of online information, recommender systems have been pervasively used in real-world business services and widely studied in literature ~\cite{resnick1997recommender,ricci2011introduction}. Upon classical static settings, in real-world applications, data are often grown in a streaming fashion and evolving with time. For example, Snapchat users share over 400 million snaps~\cite{statista_snap} and Facebook users upload 300 million photos per day~\cite{statista_facebook}. The ever-growing data volume along with rapidly evolved data properties puts the demand of time aware and online recommender systems, which could incorporate the temporal information to handle the data temporality and update in a streaming manner to alleviate the burden of data complexity.

Deep learning techniques have been widely conducted in exploiting temporal dynamics to improve the recommendation performance~\cite{zhang2017deep, hidasi2015session, wu2017recurrent, beutel2018latent}. Despite the prominence shown recently in deep recommender systems~\cite{gong2016hashtag,van2013deep, hidasi2015session}, deep frameworks also have their own limitations. One of the well-known facts is that deep recommender systems are usually deterministic approaches, which only output point estimations without taking the uncertainty into account. It significantly limits their power in modeling the randomness of the measurement noises~\cite{shi2017zhusuan} and providing predictions of the missing or unobserved interactions in recommender systems. As probabilistic approaches, especially Bayesian methods, provide solid mathematical tools for coping with the randomness and uncertainty, it motivates us to conduct streaming recommendations from the view of Deep Bayesian Learning (DBL) to conjoin the advantages of probabilistic models and deep learning models. Though some recent attempts have been made on integrating probabilistic approaches with deep autoencoder architecture for recommendation tasks~\cite{gupta2018hybrid,li2017collaborative,sachdeva2019sequential}, they are still underpinned the static recommendation setting, which allows them to be retrospective to all the historical data during the updates.

Simply applying DBL to streaming recommendations is a non-trivial task due to the following challenges. First, coordinating the temporal dynamics is difficult given the continuous-time discrete-event recommendation process along with the protean patterns on both user and item modes. A user's preference on certain items may evolve rapidly, while on others maintaining a long-term fix. Second, the high velocity of streaming data requires an updatable model, which could expeditiously extract the prior knowledge from former time steps and effectively digest it for current predictions. Also, since the data occurrence is, in fact, continuous-valued, taking the continues time information into consideration could be potentially helpful for the knowledge distillation~\cite{beutel2018latent}. Third, the DBL frameworks are usually expensive in terms of both time and space complexities. Existing optimization algorithms often require a huge amount of computation to infer and update especially under streaming setting such as Sequential Monte Carlo, which is usually infeasible for large-scale recommendations.

To tackle the aforementioned challenges, in this paper, we propose to investigate the ways to conduct streaming recommendation by leveraging the advantages of both deep models and probabilistic processes. We stick to the factorization-based approaches due to their popularity and superiority among all collaborative filtering techniques~\cite{he2017neural}.  Specifically, we study: (1) How to model the streaming recommender system with an updatable probabilistic process?
(2) How to incorporate deep architectures into the probabilistic framework? 
(3) How to efficiently learn and update the joint framework with streaming Bayesian inference?
Through answering these three questions, we propose a \underline{C}oupled \underline{V}ariational \underline{R}ecurrent \underline{C}ollaborative \underline{F}iltering (CVRCF) framework. CVRCF incorporates deep architectures into the traditional factorization-based model and encodes temporal relationships with a coupled variational gated recurrent network, which is optimized through sequential variational inference. The main contributions are summarized as follows:

\begin{itemize}
\item Propose a novel streaming recommender system CVRCF, which incorporates deep models and the general probabilistic framework for streaming recommendations;
\item Build up a linkage between probabilistic process and deep factorization based model under a streaming setting with sequential variational inference leveraging a continues-time discrete-event cross RNN model;
\item Empirically validate the effectiveness of CVRCF on different real-world datasets comparing with the state-of-the-art, explore the temporal drifting patterns learned from CVRCF, and analyze the model sensitivities.
\end{itemize}

\section{Preliminaries}

\textbf{Notations}: Before discussing the proposed framework CVRCF for streaming recommendations,  we first introduce the mathematical notations.  We consider the streaming interactions as a continues-time discrete-event process. Equipped with this viewpoint, we denote $T \in \mathbb{N}$ as the discrete time step and the inputs of a streaming recommender system can be denoted as a list of user-item interactions $\{x_{ij}^T\}$ with their occurrence time $\{\tau_{ij}^T\}$, where $x_{ij}^T$  denotes the interaction event of the $i^\text{th}$ user and the $j^\text{th}$ item occurred between time step $T-1$ and $T$, $\tau_{ij}^T$ denotes the concrete time that $x_{ij}^T$ occurs. The time interval between two consecutive time steps is called granularity, which does not need to be fixed in practice.  All interactions arrived before the $T^\text{th}$ time step are denoted as $\{ x_{ij}^{\le T} \}$ (or $\{ x^{\le T} \}$).  Without loss of generality, interactions are regarded as ratings throughout this paper. 

\noindent\textbf{Problem Statement:} Based on these notations, the streaming recommendation problem we studied in this paper is defined as: for any $T=1,2,\ldots$, given the sequence of historical user-item interactions $\{ x_{ij}^{\le T-1}\}$, with the actual time information $\{\tau_{ij}^{\le T-1}\}$, we aim at predicting the upcoming interactions $\{x_{ij}^{T}\}$ in a streaming manner. The streaming manner here means that the model should be streamingly updatable. In another word, if we assume a model is achieved at $T=k-1$, then at time $T=k$, the model should be able to update based only on the data acquired between time $T=k-1$ and $T=k$, \emph{i.e.}, $\{x_{ij}^{k}\}$.

\section{Coupled Variational Recurrent Collaborative Filtering}

The core of CVRCF is a dynamic probabilistic factor-based model that consists of four components.  The first two formulate the user-item interactions and temporal dynamics, respectively.  Each of them incorporates a probabilistic skeleton induced by deep architectures.  The third component is a sequential variational inference algorithm, which provides an efficient optimization scheme for streaming updates.  The last component allows us to generate rating predictions based on the up-to-date model.




\subsection{Interaction Network}
\label{sec:uii}
Factor-based models are widely adopted in recommendation modelings.  They have shown a great success in multiple recommendation tasks~\cite{mnih2008probabilistic}.  Most of them follow the traditional matrix factorization setting, in which users and items are modeled as latent factors; and their interactions are defined as the linear combinations of these factors.  However, such simple linear combinations are often insufficient to model complex user-item interactions~\cite{he2017neural}.  Thus, we consider a deep probabilistic matrix factorization setting as follow
\begin{equation}
\small
\label{equ:inter}
x_{ij}^T|\mathbf{u}_i^T,\mathbf{v}_j^T,\sigma_{i,j,T}^2\sim\mathcal{N}(f_1(\mathbf{u}_i^T,\mathbf{v}_j^T), f_2(\mathbf{u}_i^T,\mathbf{v}_j^T,\sigma^2_{i,j,T})),
\end{equation}
where both $f_1(\cdot)$ and $f_2(\cdot)$ are represented by deep neural networks.  We represent the latent vectors of user $i$ and  item $j$ at time step $T$ as  $\mathbf{u}_i^T$ and $\mathbf{v}_j^T$, respectively.  The rating $x_{ij}^T$ is modeled as a Gaussian random variable whose location and scale values are the output of the deep networks. The environmental noise $\sigma_{i,j,T}^2$ could either be predefined as a hyperparameter~\cite{mnih2008probabilistic} or jointly learned.  It is worth pointing out that we assume the variance of $x_{ij}^T$ depends on both the latent vectors and the environmental noises, which is slightly different from the conventional probabilistic setting \cite{mnih2008probabilistic}. 


\subsection{Temporal Drifting Process}
\label{sec:td}


\begin{figure}
\begin{center}
\includegraphics[width=0.4\textwidth]{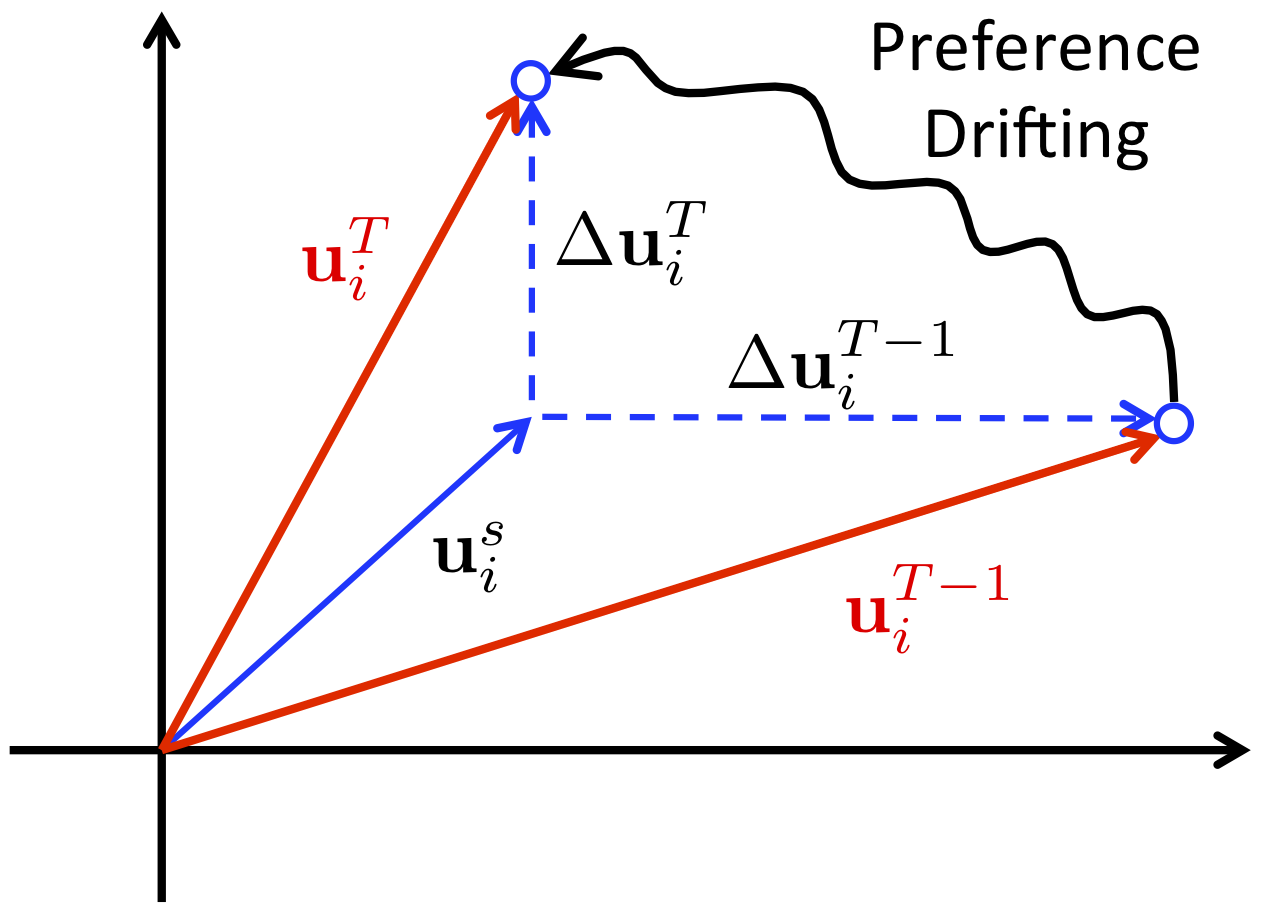} 
\end{center}
\caption{\label{fig:latent} Temporal Drifting of the a User's Latent Factor on Two Consecutive time steps.}
\end{figure} 

The temporal dynamics of a recommender system depend on the drifting of users' preferences and item popularities~\cite{rendle2008online,diaz2012real}.  A user's tastes for a certain type of items may change over time while the popularity of an item may also vary with time goes by.  To capture the inherent dynamics, we intend to encode the drifting processes into user and item latent factors based on three hypotheses: 
\begin{itemize}
\item We assume the latent factors of both user and item can be decomposed as the combination of a stationary term ($\mathbf{u}^s_i$) and a dynamic term ($\Delta \mathbf{u}_i^T$)~\cite{wu2017recurrent}.  The stationary factor captures the long-term preference, which varies slowly over time. The dynamic factor encodes the short-term changes, which evolves rapidly. An illustrative example is shown in Figure~\ref{fig:latent}, where a user's dynamic factor evolves between two consecutive time steps, causing his preference drifted from $\mathbf{u}_i^{T-1}$ to $\mathbf{u}_i^{T}$. We assume the two factors are independent of each other for simplicity. 
\item The dynamic factors of a user or an item follows a Markov process~\cite{chang2017streaming}.  The intuition of using a Markov process comes from the observation that the changing of a user's current preference could be highly affected by his former preference. 


\item The changing of latent factors of a particular user $i$ (or item $j$) between two consecutive time steps $T-1$ and $T$ depends on the time interval between the last events before these two time steps, which involves this user (or item), \emph{i.e.}, $\Delta \tau_{u,i}^{T} = \tau_{u,i}^{T}-\tau_{u,i}^{T-1}$, where $\tau_{u,i}^{T-1}$ and $\tau_{u,i}^{T}$ denote the actual time of the two last interactions of user $i$ before time step $T-1$ and $T$, respectively.  Intuitively, the longer the interval is, the larger the drifting may happen. $ \tau_{u,i}^{T}$ is defined to be equal to $\tau_{u,i}^{T-1}$ if no interactions happens between time step $T-1$ and $T$.
\end{itemize}

Upon these hypotheses, we model the evolution of hidden topics of a user (or an item), via spatiotemporal Gaussian priors, which is mathematically formulated as follows:
\begin{equation}\label{equ:prior} 
\begin{aligned}
\begin{cases}
\mathbf{u}_i^T=\mathbf{u}_i^s+ \Delta\mathbf{u}_i^T,\\ \\ 
\mathbf{u}_i^s  \sim \mathcal{N}(\mathbf{0}, \sigma^2_U\mathbf{I}),\\ \\
\Delta\mathbf{u}_i^T|\Delta\mathbf{u}_i^{T-1} \sim \mathcal{N}(\pmb{\mu}_{u,i,T}, \pmb{\Sigma}_{u,i,T}).
\end{cases}
\end{aligned}
\end{equation}

It is worth pointing out that only the users, which have interactions between time $T$ and $T-1$, need to be considered here while factors of users who do not have interactions are assumed to be unchanged till their next interaction happens. We place the zero-mean spherical Gaussian prior on the stationary factors~\cite{mnih2008probabilistic}, where $\sigma_U$ denotes the scale hyperparameter. For dynamic factors, the kernel matrix $\pmb{\Sigma}_{u,i,T}$ is defined as a diagonal matrix here for simplicity, \emph{i.e.}, $\pmb{\Sigma}_{u,i,T}  \triangleq   diag(\pmb{\sigma}_{u,i,T}^2)$. Motivated by the recent advances in deep kernel learning, which combines the non-parametric flexibility of kernel approaches with the structural properties of deep architectures~\cite{wilson2016deep}, we further define the kernel as an output of a deep neural network $f_3(\cdot)$ to enhance its generality, \emph{i.e.}, $\pmb{\sigma}_{u,i,T}^2=f_3(\Delta\mathbf{u}_i^{T-1},\Delta \tau^T_{u,i})$. 

Coping with the last two hypotheses, this spatiotemporal kernel takes $\Delta\mathbf{u}_i^{T-1}$, which represents the user's dynamic preference at last time step, as a spatial effect to decide the drifting uncertainty and it is stationary for temporal effect, which means $\pmb{\Sigma}_{u,i,T}$ depends on the time internal $\Delta \tau_{u, i}^T$ rather than the concrete time $\tau_{u,i}^T$ and $\tau_{u,i}^{T-1}$. For a more unified representation, we can further define $\pmb{\mu}_{u,i,T}=f_4(\Delta\mathbf{u}_i^{T-1},\Delta \tau^T_{u,i})$, where $f_4(\cdot)$ denotes a predefined deep neural network. The definition of the whole drifting prior obeys the Markov property for the discrete events on the continues timeline, which implies that the current state depends only on the former state. It is also applicable to employ other state dependency correlations and network structures. Similar prior with corresponding notations is defined for items.

\begin{figure*}[t!]
\centering
\subfigure[CVRCF sequential variational inference network.]{
\includegraphics[height=6cm]{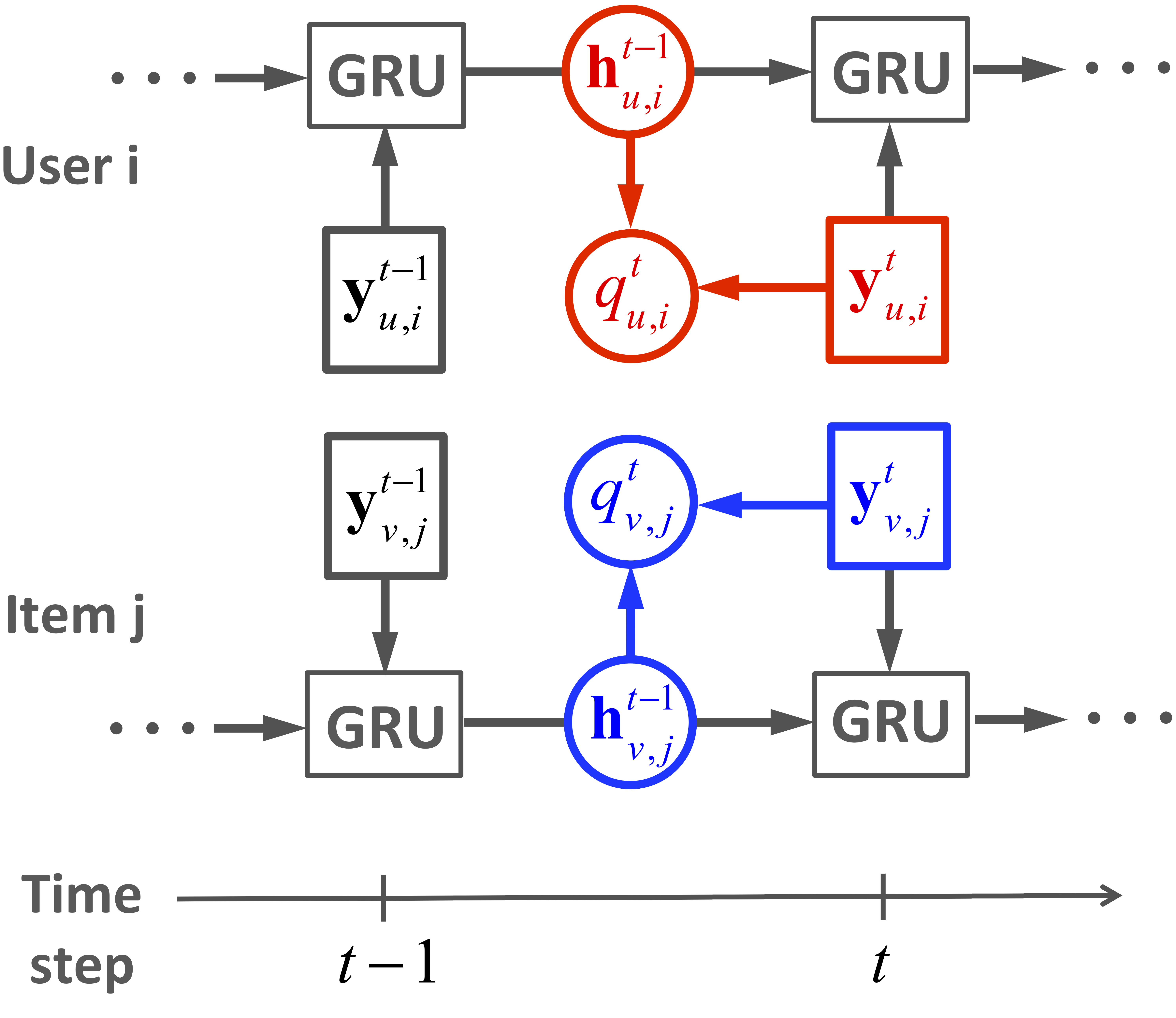}  
    \label{fig:infer}
}
\hspace{20pt}
\subfigure[CVRCF prediction network.]{
\includegraphics[height=6.1cm]{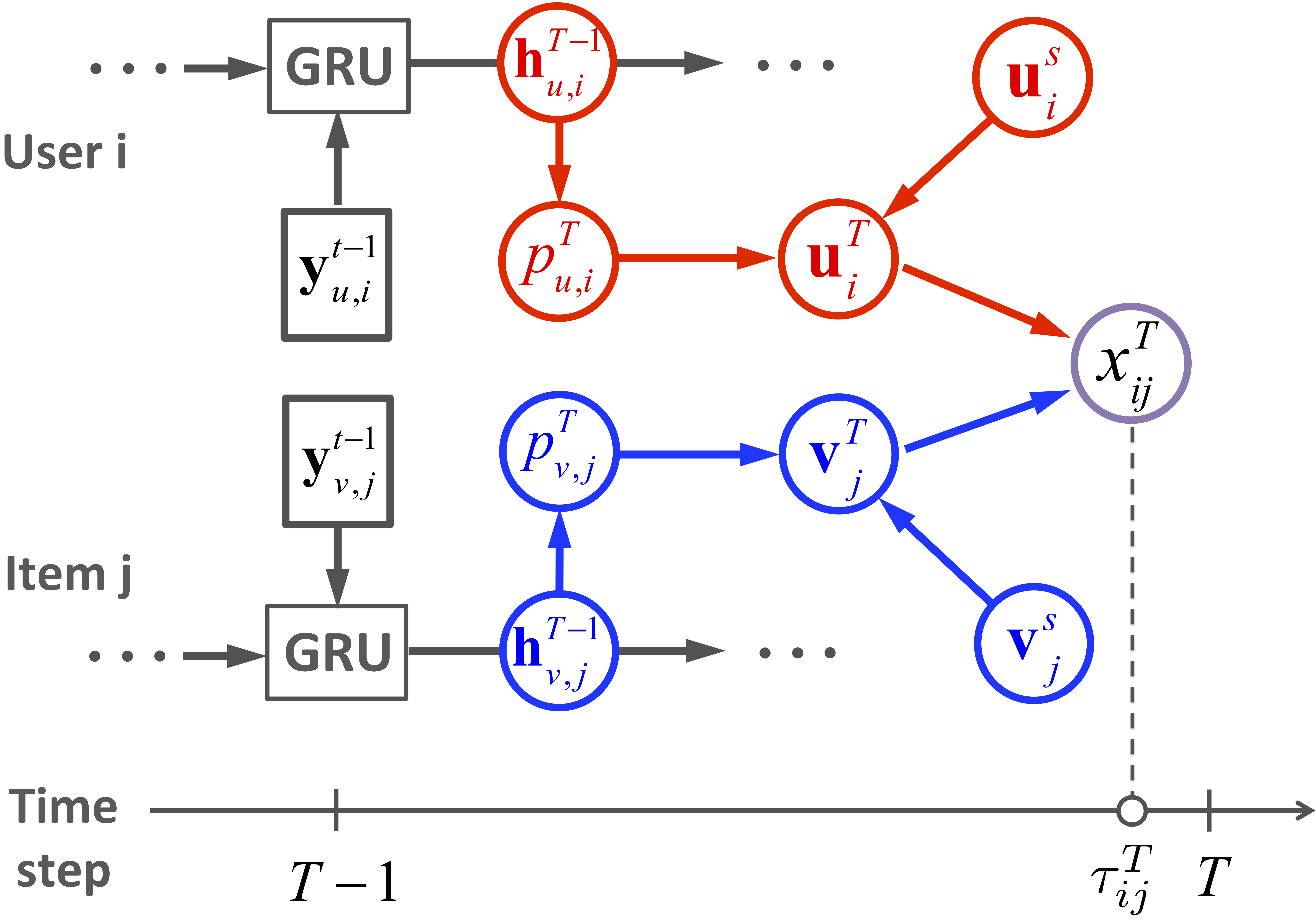} 
    \label{fig:pred}
}
\caption{\label{fig:cVRCF} Illustration of the inference and prediction networks of CVRCF. }
\end{figure*}

\subsection{Deep Sequential Variational Inference}
\label{sec:dsvi}
The third component of the CVRCF framework is the inference model.  It composites the two former components with a sequential Bayesian skeleton and associates them with the last prediction component for streaming recommendations.  

\subsubsection{Joint Distribution}
The joint distribution of all observations up to time $T$ and the latent factors is defined as follows:
\begin{equation}
\begin{aligned}
& p(x^{\le T}, \mathbf{U}^{\le T}, \mathbf{V}^{\le T}) = p(x^{\le T}, \mathbf{U}^{s}, \mathbf{V}^{s},\Delta\mathbf{U}^{\le T}, \Delta\mathbf{V}^{\le T})\\
= ~ & p(x^{\le T}, \Delta\mathbf{U}^{\le T}, \Delta\mathbf{V}^{\le T}|\mathbf{U}^{s}, \mathbf{V}^{s})p(\mathbf{U}^{s})p( \mathbf{V}^{s})\\
= ~ & p(\mathbf{U}^{s})p( \mathbf{V}^{s}) \Big[ \prod_{t\le T} p(x^t|x^{<t},\Delta\mathbf{U}^{\le t}, \Delta\mathbf{V}^{\le t}, \mathbf{U}^{s}, \mathbf{V}^{s}) \\
& \qquad\quad \times p(\Delta\mathbf{U}^{t}, \Delta\mathbf{V}^{t}|x^{<t},\Delta\mathbf{U}^{<t}, \Delta\mathbf{V}^{<t},\mathbf{U}^{s}, \mathbf{V}^{s})\Big],
\end{aligned}
\end{equation}
where $\mathbf{U}$ and $\mathbf{V}$ are the matrices of the latent factors for existing users and items. 

Our goal is to infer the posterior distribution of latent factors for every $t$, \emph{i.e.}, $p(\mathbf{U}^{t},\mathbf{V}^{t}|x^{\le t}), \forall t\le T$.   However, it is intractable for direct inferences based on the current model assumptions.  To overcome this challenge, existing works usually focus on two types of approaches - Sequential Monte Carlo methods (SMC) \cite{doucet2001introduction} and Variational Inference methods (VI) \cite{broderick2013streaming}.  The traditional sequential Bayesian updating usually uses SMC methods (\emph{a.k.a.}, particle filtering) to deal with intractable target posterior distributions.  Although this approach is very accurate when suitable proposal distributions and enough particle samples are presented, the sampling process is often too slow to apply to high dimensional and large-scale data~\cite{saeedi2017variational}.  On the other hands, the variational inference is much faster compared to SMC.  However, the accuracy highly depends on the approximation distribution, especially in streaming settings~\cite{turner2011two}.  Although there are hybrid models combine both algorithms together~\cite{gu2015neural,naesseth2017variational}, the computational complexity makes it prohibited for large-scale recommender systems.  To trade-off the model scalability and accuracy, we consider the streaming variational inference framework~\cite{broderick2013streaming} by leveraging deep neural networks as the variational approximator to obtain more flexible posteriors.

\subsubsection{Sequential Variational Inference Network}
\label{sec:dvap}




Before introducing the deep architectures, we first assume the latent factors can be partitioned into independent units followed by the traditional mean-field approximation: 
\begin{equation}
\label{equ:vap}
\begin{aligned}
&q(\Delta\mathbf{U}^{\le T}, \Delta\mathbf{V}^{\le T}| x^{\le T},\mathbf{U}^{s},\mathbf{V}^{s}) = q(\Delta\mathbf{U}^{\le T}|x^{\le T})q(\Delta\mathbf{V}^{\le T}|x^{\le T}),
\end{aligned}
\end{equation}
where $q$ denotes the approximated variational posterior. Further, each user (or item) is placed by a Gaussian variational posterior as follows:
\begin{equation}\label{equ:q}
q(\Delta \mathbf{u}_i^t| \Delta \mathbf{u}_i^{\le{t-1}},x_i^{\le t}) = \mathcal{N}(\pmb{\mu}_{u,i,t}^\ast, \pmb{\Sigma}^\ast_{u,i,t}),  \forall 1\le t \le T,
\end{equation}
where $\pmb{\Sigma}_{u,i,t}$ is diagonal with the similar definition as the priors defined in Equ.~\eqref{equ:prior}. $x_i^{\le t}$ denotes all the interactions related to user $i$ before time step $t$.


To infer the variational posterior, we propose a Coupled Variational Gated Recurrent Network structure (CVGRN) leveraging two variational Gated Recurrent Units (GRUs) for users and items, respectively.  \Cref{fig:infer} demonstrates the key idea of the proposed inference network. Blocks represent the inputs of two GRUs at different time steps. $q_{u,i}^{t}$ and $q_{v,j}^{t}$ represent the approximated posterior distribution $q(\Delta \mathbf{u}_i^{t}| \Delta \mathbf{u}_i^{\le t-1},x_i^{\le {t}}) $ and $q(\Delta \mathbf{v}_j^{t}| \Delta \mathbf{v}_j^{\le t-1},x_j^{\le t }) $, which are inferred based on the GRUs output states $\textbf{h}_{u,i}^{t-1}$ and $\textbf{h}_{v,j}^{t-1}$ and the interactions elated to user $i$ and item $j$ between time step $t-1$ and $t$, \emph{i.e.}, $\{x_i^{t}\}$ and $\{x_j^{t}\}$.  Specifically, assume a user and a movie interact with each other at time $t$. The red and blue blocks denote the inputs of the user chain and item chain at time step $t$, respectively, which are denoted as $\mathbf{y}^{t}_{u,i}$ and $\mathbf{y}^{t}_{v,j}$. These two inputs are constructed based on user $i$'s or item $j$'s interactions between time steps $t-1$ and $t$, respectively. For example, $\mathbf{y}_{u,i}^{t}$ is defined as $\mathbf{y}_{u,i}^{t}= [\mathbf{W}_u\cdot\mathbf{x}_{u,i}^{t}, \log(\Delta\tau_{u,i}^{t}), 1_{u, \text{new}} ]$, where $\mathbf{x}_{u,i}^{t}$ denotes a sparse vector consisting of the ratings $\{x_i^{t}\}$ given by user $i$ in time interval $\Delta\tau_{u,i}^{t}$. $\mathbf{W}_u$ is an embedding matrix, which is employed to reduce the length of GRUs inputs for alleviating intermediate data explosion. $1_{u, \text{new}}$ indicates whether a user is a new user or not~\cite{wu2017recurrent}. The log interval $\log(\Delta\tau_{u,i}^{t})$ is concatenated into the inputs to encode continues-time information~\cite{beutel2018latent}. Inferring $q_{u,i}^{t}$ is equivalent to inferring $\pmb{\mu}_{u,i,t}^\ast$ and $\pmb{\Sigma}^\ast_{u,i,t}$ in Equ.~\eqref{equ:q}, which are calculated as: $[\pmb{\mu}_{u,i,t}^\ast, \pmb{\Sigma}^\ast_{u,i,t}] = f_5 (\mathbf{h}_{u,i}^{t-1}, \mathbf{y}_{u,i}^{t})$. $f_5$ is a deep neural network.

Since all of the users (or items) share the same RNN chain, the model size could be largely reduced.  Moreover, to further reduce the number of latent variables,  the conditioned prior distributions of the dynamic factors $\Delta\mathbf{u}_i^T|\Delta\mathbf{u}_i^{T-1}$, which is defined in Equ.~\eqref{equ:prior}, are assumed to be parameterized by the latent states, \emph{i.e.},  $[\pmb{\mu}_{u,i,t}, \pmb{\Sigma}_{u,i,t}] = [f_4(\mathbf{h}_{u,i}^{t-1},\Delta \tau^T_{u,i}), f_3(\mathbf{h}_{u,i}^{t-1},\Delta \tau^T_{u,i})]$. To further encode the temporal information, we exponentially decay the latent state variables at each time step~\cite{mozer2017discrete} as $\mathbf{h}_{u,i}^{t} \leftarrow  \mathbf{h}_{u,,i}^{t} \cdot e^{ \frac{\Delta\tau_{u,i}^{t}}{\lambda} }$, where $\lambda$ is a predefined decay rate. 

\subsubsection{Objective Function}

Considering RNN as a graphical model, we leverage the conditionally independency between current latent state and future inputs, and have $\mathbf{h}^{t}\indep x^{>t} | \mathbf{h}^{t-1},x^{t}$. Then Equ.~\eqref{equ:vap} could be written as:

\begin{equation}\label{equ:vap2}
\begin{aligned}
q(& \Delta\mathbf{U}^{\le T},\Delta\mathbf{V}^{\le T}|x^{\le T},\mathbf{U}^{s},\mathbf{V}^{s})\\
&=\prod_{t\le T}q(\Delta\mathbf{U}^t|x^{\le t},\Delta\mathbf{U}^{<t})q(\Delta\mathbf{V}^t|x^{\le t},\Delta\mathbf{V}^{<t}).
\end{aligned}
\end{equation}

To obtain the objective function, we try to follow the traditional variational autoencoder to derive a variant variational lower bound. We start from the joint log likelihood and drive the objective function as follows:
\begin{equation}\small
\begin{aligned}
& \log p(x^{\le T}, \mathbf{U}^{s},\mathbf{V}^{s}) = \log p(x^{\le T}| \mathbf{U}^{s},\mathbf{V}^{s}) + \log p(\mathbf{U}^{s} )  + \log p(\mathbf{V}^{s} )  \\
& = \int  \log  p(x^{\le T},\Delta\mathbf{U}^{\le T},\Delta\mathbf{V}^{\le T} | \mathbf{U}^{s},\mathbf{V}^{s})  d \Delta\mathbf{U}^{\le T} d \Delta\mathbf{V}^{\le T}  + \log p(\mathbf{U}^{s} )  + \log p(\mathbf{V}^{s} )\\
 &   \ge \int q(\Delta\mathbf{U}^{\le T},\Delta\mathbf{V}^{\le T}|x^{\le T}) \log \frac{p(x^{\le T},\Delta\mathbf{U}^{\le T},\Delta\mathbf{V}^{\le T} | \mathbf{U}^{s},\mathbf{V}^{s} )}{q(\Delta\mathbf{U}^{\le T},\Delta\mathbf{V}^{\le T}|x^{\le T})} d \Delta\mathbf{U}^{\le T} d \Delta\mathbf{V}^{\le T}  \\ 
 & + \log p(\mathbf{U}^{s} )  + \log p(\mathbf{V}^{s} )  \\
& = \sum_{t\le T}\Big\{ 
E_{q(\Delta\mathbf{U}^t|x^{\le t},\Delta\mathbf{U}^{<t}),q(\Delta\mathbf{V}^t|x^{\le t},\Delta\mathbf{V}^{<t})} [\log p(x^{t}|x^{<t}, \mathbf{U}^{\le t}, \mathbf{V}^{\le t} )]\\
& -KL(q(\Delta\mathbf{U}^t|x^{\le t},\Delta\mathbf{U}^{<t}) || p(\Delta\mathbf{U}^{t}|\Delta\mathbf{U}^{<t})) \\
& -KL(q(\Delta\mathbf{V}^t|x^{\le t},\Delta\mathbf{V}^{<t}) || p(\Delta\mathbf{V}^{t}|\Delta\mathbf{V}^{<t})) \Big\}  \\
& + \log p(\mathbf{U}^{s} )  + \log p(\mathbf{V}^{s} ). 
\end{aligned}
\end{equation}

To further simply the expression, we denote the probabilities $q(\Delta\mathbf{U}^t|x^{\le t},\Delta\mathbf{U}^{<t})$, $q(\Delta\mathbf{V}^t|x^{\le t},\Delta\mathbf{V}^{<t})$, $p(\Delta\mathbf{U}^{t}|\Delta\mathbf{U}^{<t})$, $p(\Delta\mathbf{V}^{t}|\Delta\mathbf{V}^{<t})$, and $p(x^{t}|x^{<t},\mathbf{U}^{\le t},\mathbf{V}^{\le t} )$, as $q_u^t$, $q_v^t$, $p_u^t$, $p_v^t$, and $p_{x}^t$, respectively. Based on the former definitions, the objective function is defined as a timestep-wise variational lower bound as follows:
\begin{equation}
\begin{aligned}
\mathcal{L} = & \sum_{t\le T}\Big\{ 
\mathbb{E}_{q_u^t,q_v^t} [ \log p^t_{x}]  - \text{KL}(q_u^t || p_u^t) - \text{KL}(q_v^t|| p_v^t) \Big\} \\
& ~~ + \log p(\mathbf{U}^s) + \log p(\mathbf{V}^s).
\end{aligned}
\end{equation}
It is worth pointing out that the expectation term is calculated based on sampling, \emph{i.e.}, $ \mathbb{E}_{q_u,q_v} [\log p^t_{x}]  \simeq \frac{1}{L} \sum_{l=1}^L \log p(x^{t}|x^{<t},\mathbf{U}^{\le t,l},\mathbf{V}^{\le t,l} )$, where $L$ is the number of samples we wish to use to estimate the quantity.  We specifically set $L=1$ for every iteration in the implementation following the setting in conventional Variational Auto-Encoder~\cite{kingma2014auto} and adopt the reparameterization trick for feasible optimization.


As the rating sequence of each user or item could be infinite long under the streaming setting, which makes it infeasible to feed the whole sequences into the RNNs, this step-wise objective function allows us to truncate the sequences into multiple segmentations for a streaming inference. In another words, assume $q_u^T$, $q_v^T$, $p_u^T$, $p_v^T$, $\mathbf{U}^s$ and $\mathbf{V}^s$ are achieved at time step $T$, they could be treated as the prior distribution of the latent variables at time step $T+1$ and updated based on the new interactions $\{x_{ij}^{k}\}$, the CVRCF framework, and the following step-wise objective function:
\begin{equation}
\begin{aligned}
\mathcal{L} = & \Big\{ 
\mathbb{E}_{q_u^{T+1},q_v^{T+1}} [ \log p^{T+1}_{x}]  - \text{KL}(q_u^{T+1} || p_u^{T+1}) - \text{KL}(q_v^{T+1}|| p_v^{T+1}) \Big\} \\
& ~~ + \log p(\mathbf{U}^s) + \log p(\mathbf{V}^s).
\end{aligned}
\end{equation}

It is worth pointing that as stated in Section~\ref{sec:td}, we assume the stationary factors $\mathbf{U}^s$ and $\mathbf{V}^s$ represent long-term users' preferences and item popularities. Thus, they should also be updated at each time-step. However, they remain the same between two consecutive time steps while the dynamic factors keep evolving.



\subsection{Prediction Network}


The prediction model is based on the generation model described in \Cref{fig:pred}. At any testing time between time steps $T-1$ and $T$, to predict a specific ratings of a user $i$ to an item $j$, we first calculate the expectations of the current latent representations $\mathbf{u}_i^{T}$ and $\mathbf{v}_j^{T}$ based on the prior distributions $p_{u,i}^T$ and $p_{v,j}^T$, and the stationary factors $\mathbf{u}_i^s$ and $\mathbf{v}_j^s$. The ratings is then predicted based on the distribution parameterized by the interaction network in Equ.~\eqref{equ:inter}, \emph{i.e.},  $\mathbb{E}(x_{ij}^{T}|\cdot) =f_1(\mathbb{E}(\mathbf{u}_i^{T} ), \mathbb{E}(\mathbf{v}_j^{T}))$.  Similarly, the variance could also be predicted as: $V(x_{ij}^{T}|\cdot)= f_2(\mathbb{E}(\mathbf{u}_i^{T}),\mathbb{E}(\mathbf{v}_j^{T}), \sigma^2_{i,j,T}  )$. $\sigma^2_{i,j,T}$ is assumed to be learnable as a function of the hidden states $\mathbf{h}_{u,i}^{T-1}$ and $\mathbf{h}_{v,j}^{T-1}$ in our implementation.

\section{Experiments}

 \begin{table}[t!]
  \caption{Dataset statistics.}
\centering
\setlength{\tabcolsep}{4pt}
\begin{tabular}[0.1\textwidth]{ccccc}\toprule
& \# of User & \# of Items  & Time Spanning & Granularities\\ 
    \hline
    MT  & $53,275$      & $30,686$        &  $2013 \sim 2018$  & 4 Weeks  \\
    ML-10M & $71,567$ & $10,681$          &  $1995 \sim 2009$ & 2 Weeks  \\
        Netflix & $480,189$  & $17,700$    &   $1999 \sim 2006$  & 2 Weeks  \\
\bottomrule
\end{tabular}
 \vspace{-10pt}
\label{tab:datasets}
  \end{table}

     
In this section, we empirically evaluate the performance of CVRCF framework by analyzing three major aspects. 
\textbf{Q1}: What are the general performance of CVRCF compared with the other baselines? \textbf{Q2}: What are the temporal drifting dynamics of users and items we could learned? \textbf{Q3}: What are the sensitivities of the model to the key hyperparameters? The code of CVRCF is available at GitHub: {\color{blue}\url{https://github.com/song3134/CVRCF}}.

\subsection{Datasets}
Three widely-adopted benchmark datasets shown in Figure~\ref{tab:datasets} are employed in our experiments. Detailed statistics of them are elaborated as follows:

\begin{itemize}
\item \textbf{MovieTweetings (MT) \cite{MT}}: It is a benchmark dataset consisting of movies ratings that were contained in well-structured tweets on Twitter. It contains $696,531$ ratings ($0$-$10$) provided by $53,275$ users to $30,686$ movies. All ratings are time-associated spanning from $02/28/2013$ to $04/07/2018$. The granularity is defined as four weeks. 

\item\textbf{MovieLens-10M (ML-10M) \cite{ML}}: It contains ten million ratings to $10,681$ movies by $71,567$ users spanning from $1995$ to $2009$. The granularity is defined as four weeks.

\item \textbf{Netflix \cite{NF}}: The Netflix challenge dataset consists of $100$ million ratings by $480,189$ users to $17,700$ movies from $1999$ to $2006$. The granularity is defined as two weeks.
\end{itemize}


%
%
%

\begin{table}\small

\centering
\caption{An overview of all experimental methods.} 
\hspace{4pt}
\setlength{\tabcolsep}{3pt}
\begin{tabular}{c c c c c c} 
\toprule 
& \multirow{2}{*}{\diagbox[width=2.7cm, height=0.7cm]{Methods}{ Categories}} & \multirow{2}{*}{Streaming} &  Temporal &  \multirow{2}{*}{Probabilistic} &  \multirow{2}{*}{Deep}   \\ 
& & & Involved & & \\
\midrule 
& PMF               & 		        &                     & \checkmark &   \\  
&  time-SVD++   &  		       & \checkmark &  		   &  \\ 

&  sD-PMF           & \checkmark &  		     &\checkmark  & 		\checkmark 	 \\ 
& sRRN                     & \checkmark   & 	\checkmark 	     &      & \checkmark \\ 

& sRec              &  \checkmark  & \checkmark  &\checkmark  & \\
& {\color{blue} CVRCF (proposed)}        &  \checkmark  & \checkmark  &\checkmark  & \checkmark \\

\bottomrule 
\end{tabular}
\label{tab:baseline} 
\end{table}

\subsection{Baselines}
As our main focus is factorization-based approaches, five representative factorization-based baseline algorithms, including two batch algorithms and three streaming algorithms are selected for comparison from different perspectives shown in Table~\ref{tab:baseline}. Brief descriptions of these methods are listed as follows.


\begin{itemize}
\item\textbf{PMF}~\cite{mnih2008probabilistic}:
Probabilistic Matrix Factorization is a conventional recommendation algorithm, which does not consider temporal information. 

\item\textbf{TimeSVD++}~\cite{koren2010collaborative}: The temporal-envoled variation of the classical static factor-based algorithm SVD++. We implement it with Graphchi~\cite{kyrola2012graphchi} C++ pacakge.

\item\textbf{sD-PMF}: A streaming version of the PMF model combined with the deep interaction network, which is employed in the CVRCF Framework.  This model is used to test the effectiveness of the dynamic factors optimized with the RNN structure in CVRCF.

\item \textbf{sRec}~\cite{chang2017streaming}: Streaming Recommender System is the state-of-the-art shallow dynamic recommendation model. It is a probabilistic factor-based model optimized with a recursive mean-field approximation.

\item\textbf{sRRN}~\cite{wu2017recurrent}:
A streaming variation of Recurrent Recommender Network (RRN), which is a state-of-the-art deep heuristic streaming recommendation model. 
\end{itemize}


\subsection{Experimental Setup}

For each dataset, we segment the data along timeline into three parts with ratios $4:1:5$ serving as training, validation, and testing sets, respectively. 

\subsubsection{Training Settings} During the training phase, the training and validation sets serve as the historical datasets to decide the best hyperparameters for all methods. As each user or movie may have too many ratings, to reduce and memory and protect the feasible use of GRU structures, we truncate the training sequences along the timeline into batches for the user and movie chain, respectively.  This will affect the RNN effectiveness to some extent, but by varying the number of training epoch, it does not have an obvious influence on the experimental results during our experiments.  Moreover, to protect the stationary factor get faster trained, in each epoch, every truncated batch is processed with multiple iterations.  The number of this iteration hyperparameter used in the training phase is set based on validation and will be further analyzed in hyperparameter analysis section.  

\subsubsection{Testing Settings}  During the testing phase, at each time step $t$, the testing is first done to get the prediction of the upcoming ratings $\{ x^{t+1} \}$, and then these ratings are assumed to arrive and be used to update the models. Different from dynamic methods, at each update, the static methods are reconstructed from scratch using all the previously arrived testing ratings including the training ratings, while the streaming models only employ the current-step arrived ratings for the current update.  Based on this setting, no later data is used to predict any former data and no temporal overlapping is existed between each pair of testing intervals. Besides, for fair comparisons, at each testing step, only ratings for existing users and items are used for testing since some baselines (e.g., PMF and time-SVD++) cannot explicitly cope with new users and items. All the experimental results are the arithmetic average of ten different times runs to ensure the reliability. The performance is evaluated via the root mean square error (RMSE).

\setlength{\tabcolsep}{7pt}
 \begin{table}
 \centering 
 \caption{The RMSE results on the three datasets.} 
\begin{tabular}{l c c c c }
\toprule 
& & \multicolumn{3}{c}{Datasets} \\ 
\cmidrule(l){3-5} 
& Methods & MT& ML-10M & Netflix \\ 
\midrule 

\multirow{2}*{Batch}  &PMF & $1.5723$  & $0.8202$ & $0.9421$ \\ 
& time-SVD++ & $1.4630$  & $0.7985$  & $0.9311$ \\ \hline 

\multirow{4}*{Streaming}   
 &sD-PMF & $1.6170$  &  $0.9017$ & $0.9992$ \\ 

&sRRN & $1.5646$ & $ 0.8003$ & $0.9236$ \\ 
& sRec & $1.4831$ & $0.8121$ & $0.9288$ \\ 
&\textbf{CVRCF} & $\mathbf{1.4567}$ & $\mathbf{0.7831}$ & $\mathbf{0.9050}$ \\ 
\bottomrule 
\end{tabular}
\label{tab:RMSE} 
\end{table}

\subsubsection{Parameter Setting}\label{sec:ps}

Settings of the hyperparameters for all the baselines follow the original papers, which result in their best performance. Hyperparameters in all the methods are selected based on cross-validation using the training and validation sets. For the static baselines PMF and timeSVD++, all of their regularization parameters are chosen over  $\{10^{-4},10^{-3},\ldots, 10^2\}$ and the sizes of their latent factors are chosen over $\{20,40, 60, 80, 100\}$.  For streaming methods, the size of the stationary factors for sRRN and CVRCF are chosen to be $20$ for all the datasets. The stationary factors for sD-PMF is chosen over $\{20, 40, 60, 80, 100\}$. The size of the dynamic factors of CVRCF is chosen to be 40 including the sizes of both mean and variance parameters. The size of the dynamic factors and the length of the RNN inputs for sRRN is chosen to be the same as CVRCF for fair comparisons. The size of the latent states ($\mathbf{h}_u$ \& $\mathbf{h}_v$) of CVRCF is set to be $20$ which is half of the length we used in sRRN.  The exponential decay factors are set to be $1$ week and $4$ weeks for the user RNN and movie RNN, respectively.  In the training phase, the truncation hyperparameters of all the RNN-based models are set to be $20$, $20$, and $10$ weeks for the three datasets, respectively, to alleviate the intermediate data explosion.

\subsection{General Evaluation  Results}

  \begin{figure}
\subfigure[Testing RMSE changing curve on MovieLens-10M dataset.]{
\hspace{-0.3cm}
\includegraphics[width=.48\textwidth]{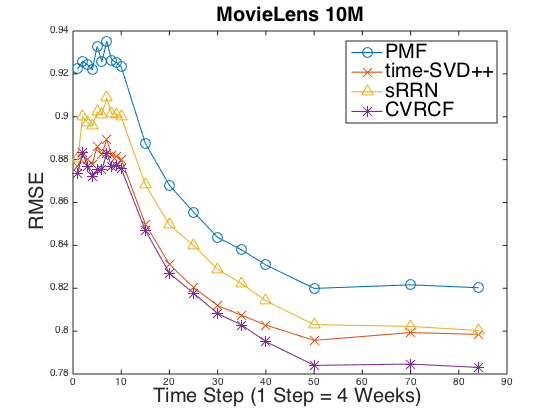}
    \label{fig:ml10M}
}
\hfill
\subfigure[Testing RMSE changing curve on netflix dataset.]{
\hspace{-0.3cm}
\includegraphics[width=.48\textwidth]{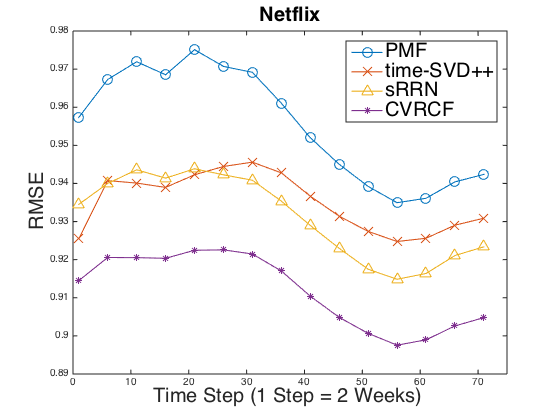}
    \label{fig:netflix}
}
\vspace{-0.2cm}
\caption{Testing RMSE changing curve of four representative methods on ML-10M and Netflix datasets.} 
\label{fig:RMSE} 
\vspace{-0.5cm}
\end{figure}

We first analyze the general performance of CVRCF model by comparing it with different categories of baselines based on the RMSE results shown in Table~\ref{tab:RMSE} and Figure~\ref{fig:RMSE}. From Table~\ref{tab:RMSE}, three conclusions could be drawn as follows. First, CVRCF outperforms all baselines on all datasets. Although time-SVD++ could achieve comparable performance on MT and ML-10M dataset, it has to be reconstructed from scratch using all of the historical data at each update. Second, CVRCF highly outperforms sD-PMF, which confirms the effectiveness of the dynamic factors employed in CVRCF for capturing the temporal relationships during the streaming process. Third, comparing with shallow probabilistic model sRec, CVRCF displays prominent improvement, which demonstrate the effectiveness of deep architectures in modeling complex drifting interactions. 

To further analyze the time-varying pattern of each method and their performance consistency on different datasets, we display the RMSE changing curves of the four representative methods on two larger datasets ML-10M and Netflix in Figure~\ref{fig:RMSE}. From the figure, we could observe that on each dataset the performance of all methods shows similar varying patterns and starting form the first testing step, CVRCF consistently achieves the best performance across two datasets with the evolving of the system. Since MovieLens-10M has the longest testing timeline among all three testing datasets, Figure~\ref{fig:ml10M} illustrates that CVRCF has stable effectiveness on the dataset with strong temporal relationships in long-term evaluation. By comparison, Netflix is a much larger dataset in terms of users and interactions. Results in Figure~\ref{fig:netflix} confirms the superiority of the proposed method on large-scale datasets. Finally, as sRRN could be treated as an ablation method of CVRCF without the probabilistic component, the relative improvement of the proposed method on the general performance validates the effectiveness of combining probabilistic approach in capturing the prospective process of streaming data generation.





\subsection{Evaluation of Temporal Dynamics}



%

To analyze the temporal drifting dynamics learned from CVRCF, we visualize the learned latent factors including the location factors ($\mathbf{u}_i^T$ and $\mathbf{v}_j^T$) and uncertainty factors ($\pmb{\sigma}_{u,i,T}$ and $\pmb{\sigma}_{v,j,T}$). We conduct exploration on the ML-10M dataset and update the models every half a year during testing.


\subsubsection{Drifting of the Location Factors}
We first visualize the drifting of the average location factors $\mathbf{u}_i^T$ and $\mathbf{v}_j^T$ with heatmap shown in \Cref{subfig:loc}. The X-axis denotes the index of the latent factors and the Y-axis denotes the timeline. Each factor is adjusted with centralization for joint visualization. 
From the figure, we could discover that the users' preference factors change more smoothly than movies' popularity factors, which display a block-wise changing patterns. As we update the model every half a year, the stationary factors of movies especially for the new movies are only updated or learned every half a year, which is consistent with the length of the blocks. Thus, the block-wise structure, which appears only on the movie factors, could be explained as: the movie drifting is more likely to be captured by the stationary factors, while the drifting pattern of the users is more likely to be captured by dynamic factors. Since the dynamic factors and stationary factors are defined to capture the short-term and long-term preference, respectively, the finding is also consistent with the fact that users preference usually change more frequently compared to movie popularities.


\subsubsection{Drifting of the Uncertainty Factors}
\begin{figure}[t]
\vspace{-0.2cm}
\subfigure[Drifting of average location factors of users \& movies.]{
\includegraphics[width=.25\textwidth]{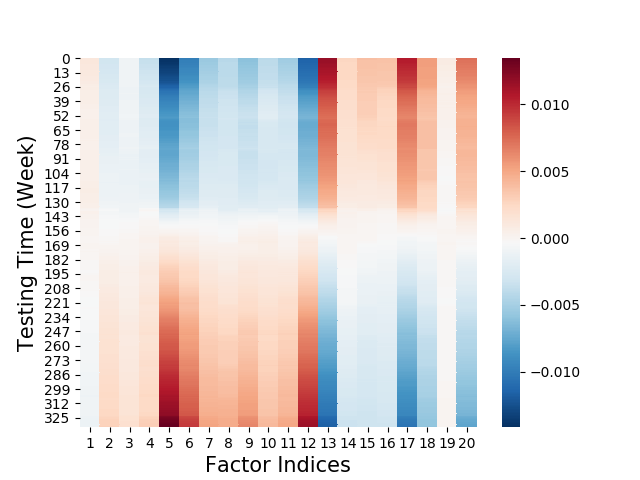}  
\includegraphics[width=.25\textwidth]{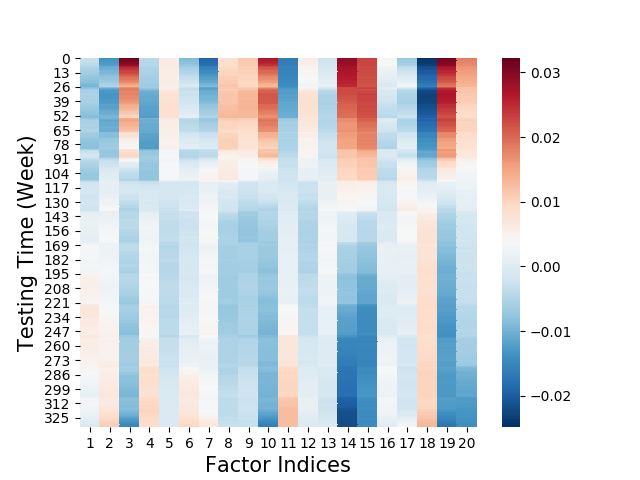} 
    \label{subfig:loc}
}
\hfill
\subfigure[Drifting of variance factors of users \& movies.]{
\includegraphics[width=.25\textwidth]{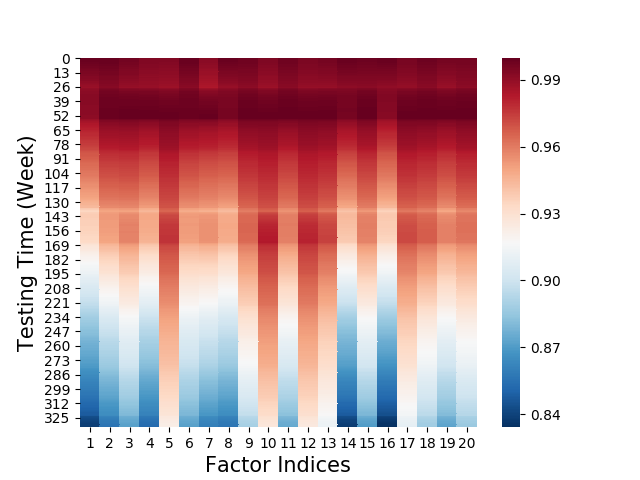} 
\includegraphics[width=.25\textwidth]{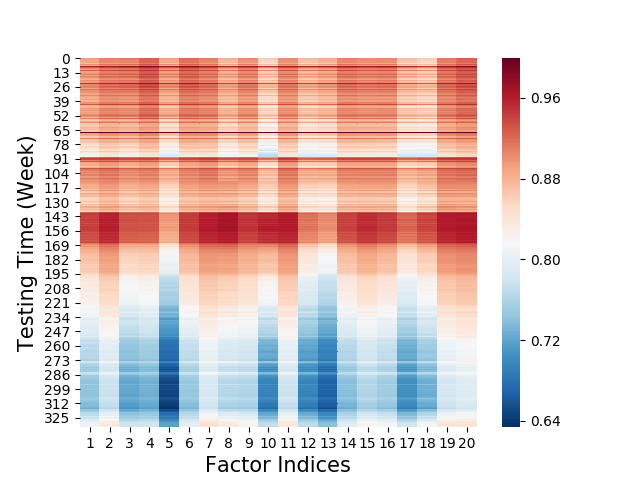} 
    \label{subfig:var}
}
\vspace{-0.3cm}
\caption{Drifting of average latent factors learned from CVRCF on the ML-10M dataset.} 
\label{fig:drifting} 
\vspace{-0.3cm}
\end{figure}

\Cref{subfig:var} displays the drifting of the average uncertainty factors learned from CVRCF. Each column is first normalized with $L_{\infty}$-norm. There are two major observations we could find from~\Cref{subfig:var}.  From an overall perspective, with the evolving of the system, the variances of the learned dynamic factors decrease.  This is because the incremental ratings provide more information for each user and item, and reduce the uncertainties of the whole system during the testing phase.  From the local perspective, at some time steps, the variance of the latent factors are sharply increased and then slowly decreased. This is because, at some time steps, users and movies increase are dramatically. The cold-start problem introduced by the incremental users and items may raise the uncertainties of the system within a short time but would be alleviated with time goes by.  In other words, although new users and items are continually enrolled, the number of ratings related to them could be deficient at first and then increasing over time. 



\begin{figure*}[t]
\centering
\hspace{5pt}
\subfigure[Effects of the train epoch \& training batch iteration.]{
\includegraphics[width=.3\textwidth]{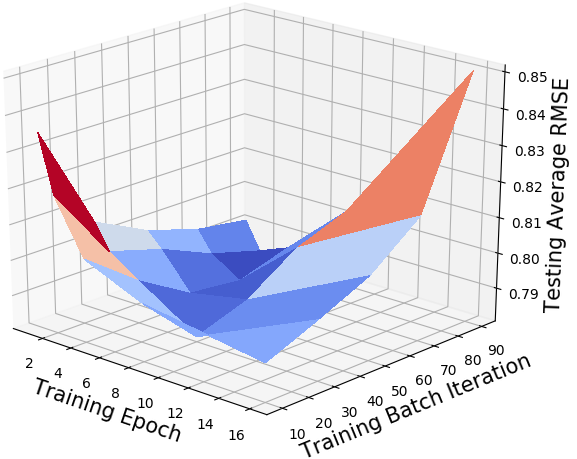}
    \label{fig:train}
}
\hspace{5pt}
\subfigure[Effects of testing batch iteration \& update interval.]{
\includegraphics[width=.3\textwidth]{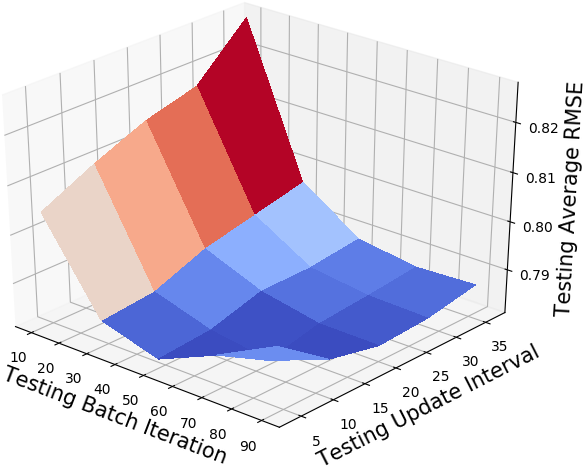}
    \label{fig:test}
}
\hspace{5pt}
\subfigure[Effects of the granularities.]{
\includegraphics[width=.3\textwidth]{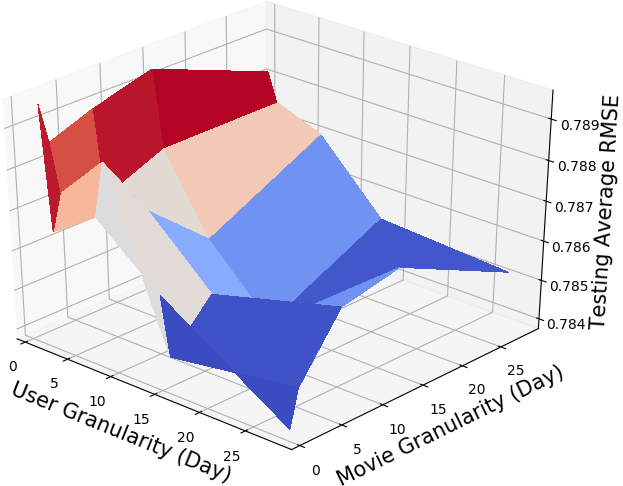}
    \label{fig:gran}
}
\vspace{-0.4cm}
\caption{Analysis of five key hyperparameters on ML-10M datasets.} 
\vspace{-0.2cm}
\label{fig:para} 
\end{figure*}
\subsection{Hyperparameter Sensitivity Analysis}
\label{sec:hyper}

Finally, we study the sensitivity of CVRCF to different hyperparameters using the ML-10M dataset. We pick five hyperparameters, which are the most influential ones in our experiments, and analyze their effects by coupling some of them.  These pairwise effects are displayed in Figure~\ref{fig:para}.

\subsubsection{Training Epochs \& Training Batch Iterations} We first analyze the pairwise effects of the training epoch and the training batch iterations.  Figure~\ref{fig:train} shows that these two parameters highly affects the learning process and may cause overfitting or underfitting when the product of them are too large or too small. With the number of training batch iteration increasing, less epoch should be adopted to protect the testing effectiveness. This may be because:  since the stationary factors are outside the RNNs and have high degrees of freedom, they may get overtrained when the batch iteration is setting too large given fixed training epoch. Thus, early stopping should be employed via limiting the number of epochs to prevent the RNN structures not further learning effectively. On the contrary, insufficient batch iterations would limit the power of stationary factors in capturing long-term preferences.


\subsubsection{Testing Batch Iterations \& Testing Update Interval} Secondly, we focus on the testing phase and analyze the influence of the testing batch iterations and the length of the model updating interval. As shown in Figure~\ref{fig:test}, for a fixed testing update interval, with the increasing of the testing batch iterations, the testing performance first decreases and then increases. This might because: in the testing phase, new ratings, users, and items never stop to arrive. Insufficient testing batch iterations would highly affect the learning of latent factors especially for the stationary factors of new users or items. On the contrary,  superfluous iterations would also lead to overfitting as in the training phase described above. Besides, with the enlarging of the testing update interval, ratings in each batch increase which requires more updating iterations under the same remaining settings.

\subsubsection{Granularities} Finally, we explore the effect of the granularities. We assume the two granularities defined for users and movies could be different for a more general treatment. From~\Cref{fig:gran}, we can see that although different granularities do affect the results, their influences are shown to be very trivial based on the scale of the $Z$-axis. Moreover, user granularity seems to have larger effects than movie granularity and its optimal value is shown to be lower than movie granularity. This may illustrate that the users' preferences are varying more frequently than the items' popularities. 

\section{Related Work}
\noindent\textbf{Streaming Recommender Systems.} Beyond traditional static settings, streaming recommender systems have attracted widespread concerns in coping with the high data velocity and their naturally incremental properties~\cite{agarwal2010fast, chang2017streaming}. Different from static time-aware models~\cite{koren2010collaborative,du2015time, wang2016coevolutionary, kapoor2015just, hosseini2017recurrent}, which only take account of temporal dynamics without updating in an streaming fashion, streaming recommender systems dynamically encode temporal information and generate response instantaneously~\cite{das2007google, song2008real, chen2013terec, song2017multi}. Some existing works focus on extending classical memory-based recommendation algorithms into online fashions to address the streaming challenges such as \cite{chandramouli2011streamrec} and \cite{subbian2016recommendations}. Besides memory-based methods, model-based methods~\cite{rendle2008online, diaz2012real, devooght2015dynamic} are becoming more and more popular in recent years, which conducts recommendation based on well-trained models rather than explicitly aggregating and prediction based on the similarity relationships.  Diaz-Aviles et al. leverage the active learning strategy to sample and maintain a delicately designed reservoir, thus providing a pairwise matrix factorization approach for streaming recommendation. Chang et al.~\cite{chang2017streaming} exploit  continuous Markov process (Brownian motion) to model the temporal drifting of users and items, which introduces a principled way to model data streams. Wang et al.~\cite{wang2018streaming} propose a streaming ranking-based framework based on Bayesian Personalized Ranking~\cite{rendle2009bpr} to address the user interest drifting as well as system overload problem. Although many recent advances based on deep neural networks especially RNNs have been made to model streaming inputs and capture the complex temporal dynamics~\cite{hidasi2015session, wu2017recurrent, beutel2018latent}, most of them overlook the causality inherited in the data generation process, which is one of the main aspects considered in our framework via the deep Bayesian learning. 


\noindent\textbf{Deep Recommender Systems.} Deep learning techniques have brought vast vitality and achieve dramatic improvement in recommender systems~\cite{zhang2017deep}. They have been adopted in various recommendation tasks as well as accommodating different data sources~\cite{van2013deep, gong2016hashtag, wu2017recurrent}. From the perspective of the general framework, deep recommender systems could be categorized into solely deep models, which conduct recommendations based only on deep frameworks~\cite{gong2016hashtag, song2016multi, he2017neural}; and integration models, which integrate deep techniques with traditional recommender systems~\cite{van2013deep, wang2016collaborative,wang2017your}. From the perspective of deep frameworks, these models could also be divided into: (1) single deep models, which are built upon single neural building blocks such as multi-layer perceptron~\cite{he2017neural}, convolutional neural network~\cite{wang2017dynamic}, and recurrent neural network~\cite{wu2017recurrent}; and (2) composite models, which are constructed with different deep learning techniques~\cite{zhang2016collaborative}. From the first viewpoint of devision, the proposed framework could be categorized as an integration model,  which combines and probabilistic recommender systems with deep learning models. It is also a hybrid deep models, which jointly incorporates RNN and MLP structures. The coupled variational inference structure also provides its' uniqueness comparing to other streaming deep recommender systems.

\section{Conclusion and Future Work}
\label{Conclusion}
In this paper, we focus on the recommendation problem under streaming setting and propose a deep streaming recommender system - CVRCF. CVRCF incorporates deep architectures into traditional factorization-based model and encodes the temporal relationship with Gaussian-Markov components. Standing upon the sequential variational inference, CVRCF is optimized leveraging a cross variational GRU network and could continually update under the streaming setting. By conducting experiments on various real-world benchmark datasets, we empirically validate the effectiveness our proposed framework, explore the learned drifting patterns, and validate the stability of our framework. Future work will center on exploring different assumptions of stochastic processes of the dynamic factors and incorporate other deep learning structures, such as graph neural networks, into the proposed framework.

\section{Acknowledgments}
\label{ack}

The authors thank the anonymous reviewers for their helpful comments. This work is, in part, supported by DARPA under grant \#W$911$NF-$16$-$1$-$0565$ and \#FA$8750$-$17$-$2$-$0116$, and NSF under grant \#IIS-$1657196$ and \#IIS-$1718840$. The views and conclusions contained in this paper are those of the authors and should not be interpreted as representing any funding agencies.



%
\bibliographystyle{ACM-Reference-Format}
\bibliography{Rec}

\end{document}